# Radio Map Estimation in Urban Environments using Deep Learning

Jonathan O'Shea, DCU School of Electronic Engineering; Dr. Conor Brennan, DCU School of Electronic Engineering

*Abstract*— **Accurate path loss prediction is a critical component of wireless network planning. Current path loss prediction methods typically struggle to balance the trade-off between accuracy and computational efficiency. This paper proposes the Efficient Attention Radio Map Estimation Network (EA-RMENet) which is an image data-driven, deep learning (DL) model designed for radio map estimation (RME). EA-RMENet uses a U-Net framework with an EfficientNetB5 encoder, Attention Gated (AG) skip connections, and Atrous Spatial Pyramid Pooling (ASPP). The EfficientNet encoder uses compound scaling to balance accuracy and efficiency. AG skip connections suppress irrelevant features, and the ASPP captures a multi-scale context. The model has a test prediction RMSE of 0.0334 on the RadioMapSeer3D dataset with an inference time of 0.022 seconds/sample. In the ICASSP 2023 Radio Map Prediction Challenge, the model ranks third with a competitive RMSE of 0.0406 this highlights the models potential for real-world RME.**

## I. Introduction

Wireless radio frequency (RF) communication systems enable essential services such as mobile connectivity and 5G services. One key element affecting wireless network performance is signal path loss. Path loss can be defined as the reduction in signal power as it propagates through an environment from transmitter (Tx) to receiver (Rx). Accurate path loss modelling is essential for effective network planning and ensuring optimal coverage. Path loss modelling in urban environments is a particularly complex challenge due to the presence of dense infrastructure, complex building layouts and varying building heights [1]. Solving these challenges requires advanced modelling techniques that can accurately capture the relevant spatial characteristics effecting path loss. Traditional path loss models can be generally categorised as Empirical or Deterministic. Empirical models such as COST 231 Hata and ECC-33 offer efficiency but lack accuracy in complex environments. Deterministic models, such as ray tracing, offer high accuracy but at the cost of significant computational and time resources [2].

Deep-Learning (DL) methods, primarily encoder-decoder architectures such as U-Net, have emerged as promising solutions for predicting path loss using image data such as radio maps. Path loss radio maps (RM) are images that capture environmental and RF signal propagation characteristics. These data characteristics make RMs effective training data for DL models applied to path loss prediction problems in urban environments. This paper proposes the integration of an EfficientNet [3] encoder in a U-Net architecture for path loss predictions in urban environments using RM data. EfficientNet B0-B7 variants are evaluated using the RadioMapSeer2D dataset [4]. Further model development and testing is conducted using the RadioMapSeer3D dataset [4]. Model optimisation focuses on enhancement techniques such as Attention Gates (AGs) for selective filtering of key spatial features and Atrous Spatial Pyramid Pooling (ASPP) for multi-scale context extraction [5]. The developed model, named *Efficient Attention Radio Map Estimation Network* (EA-RMENet), offers an accurate and computationally efficient alternative to traditional RF path loss models.

## II. Prior Work

Convolutional Neural Networks (CNNs) provide the foundation for DL due to their hierarchical feature extraction capabilities. State-of-the-art CNN architectures such as EfficientNet [3] are designed to improve accuracy, trainability, and efficiency while addressing inherent challenges such as optimising model depth and width. EfficientNet presents compound scaling as a method of optimally scaling depth, width, and resolution with maximum efficiency and fewer model parameters [3]. EfficientNetB0 is the baseline EfficientNet model with EfficientNetB1-B7 scaling up from EfficientNetB0 systematically. EfficientNetB5 has been adopted as a balanced trade-off between accuracy and efficiency in U-Net-based models [6] [7] used for indoor RF propagation and medical imaging tasks. U-Nets are generally used for semantic and instance segmentation tasks. U-Net builds on CNNs with an encoder-decoder structure and skip connections between the corresponding encoder–decoder layers to preserve high-resolution features. Researchers have widely applied U-Net models to RME. RadioUnet [1] exhibited the architecture's ability to achieve strong generalisation and accuracy on urban RME tasks. RadioUnet also introduced the RadioMapSeer2D dataset [1].

Several models have modified the U-Net model to improve prediction accuracy. PMNet [2] incorporated ResNet encoder blocks and integrated ASPP in the U-Net bottleneck to capture multi-scale context using parallel atrous convolutions with varying dilation rates to extract local and global features [8]. PMNet also applied data augmentation (DA) to improve generalisation. AG skip connections as implemented by the authors of [5] act as feature filters that highlight relevant features by suppressing irrelevant features [5] to improve RME efficiency. RadioTrans [9] utilised Grid Anchors (GA) to provide spatial encoding for pixel locations to enhance the positional representation of the environment relative to Tx [9]. While combinations of AGs, ASPP, DA, and GA with a U-Net framework have been researched in selective combinations [2] [9] [10], there is no known published work combining all discussed techniques with a U-Net architecture for RME. This work proposes a novel approach to integrating all discussed components to improve accuracy and efficiency for RME in urban scenarios.

## III. RADIOMAPSEER DATASET

The RadioMapSeer dataset [4] has a total of 56000 ray tracing simulated radio maps comprised of 700 urban maps. Each urban map has 80 corresponding Tx antenna locations. The 2D dataset represents cities as footprint maps with fixed-height buildings and antennas. The 3D dataset introduces varying building and antenna height for a realistic urban propagation scenario. Table I outlines the dataset parameters [4].

*Table I - RadioMapSeer3D dataset parameters as described in [4].*

| Parameter | Value |
|---|---|
| Map size | 256 x 256 |
| Resolution | 1 m / pixel |
| Transmitter power | 23 dBm |
| Antenna type | Isotropic |
| Height range of buildings | 6.6 - 19.8 m |
| Tx Height | 3 m above rooftop |
| Min PL in simulation | -162 dB |
| Max path gain | -75 dB |
| PL threshold | -104 dB |
| Min truncated path gain | 29 dB |
| PL threshold (analytic) | -111 dB |

## IV. PROPOSED METHOD

This section describes the proposed EA-RMENet design and implementation method.

### A. EA-RMENET Architecture Overview

EA-RMENET is a custom U-Net-based architecture that integrates an EfficientNet encoder with two key architectural enhancements: (1) *Attention-Gated (AG) skip connections and* (2) *Atrous Spatial Pyramid Pooling* (*ASPP*) *bottleneck.* AG skip connections operate as learnable feature filters, enabling the decoder to selectively focus on spatially important encoder features during upsampling and improve feature encoder to decoder fusion. The EfficientNet backbone extracts five hierarchical feature maps of spatial dimensions 128 x 128, 64 x 64, 32 x 32, 16 x 16, and 8 x 8. The final encoder output, an 8 x 8 x 2048 tensor, is input to an ASPP module. The ASPP extracts a multi-scale context at the model's bottleneck to capture local and global spatial patterns. ASPP uses 4 parallel atrous convolution branches with dilation rates 1, 2, 3, and 4 respectively plus a global pooling branch. A five-stage decoder progressively upsamples to an image of 256 x 256 dimensions. At each decoder stage, the AG skip connections merge with the corresponding encoder features. A final pointwise convolution generates the output normalised RME. The complete architectural framework is illustrated in Figure D.11 in Appendix D.

### B. Data Presentation

Image data is presented as 8-bit PNG images containing encoded physical path loss values [4]. The physical range for simulated radio maps in the dataset is outlined in Table I and [4] as truncated to be between -111 dB and -75 dB. The clipping at – 111 dB collapses to zero any pixels that carry no useful information, to mask noisy regions, and give zero weight to noise during metric computation [4]. Each map is linearly rescaled to [0, 1] for every pixel using (1).

$$f(i,j) = max\left\{0, \frac{PL(i,j) - PL_{min}}{PL_{max} - PL_{min}}\right\} [4] \quad (1)$$

Where $PL_{min} = -111\ dB$ and $PL_{max} = -75\ dB$. The normalised value $f \in [0,1]$ is multiplied by the maximum greyscale value (255) and rounded to the nearest integer to produce the encoded pixel value $\tilde{P}$.

$$\tilde{P}(i,j) = [255 \times f(i,j) + 0.5] \quad (2)$$

Encoded pixel value $\tilde{P}$ is divided by 255 in the model data loader to recover the [0,1] normalised representation. Data pre-processing with this method presents a uniform dynamic gain range across all maps, reduces PNG file size, improves numerical stability, and results in cleaner gradients during backpropagation.

### C. Input Representation

Each map model input tensor has either 14 or 16 channels, depending on the configuration. The channels are comprised of (1) a single-channel building footprint $B(i,j) \epsilon \{0,1\}$, (2) a single-channel Tx mask $T(i,j) \in \{0,1\}$, (3) 12 height slices derived from building heights, and (4) an optional two-channel GA.

#### 1) Building Height Slicing

Urban radio propagation is sensitive to building height, which can influence diffraction, reflection, and shadowing effects. To enable the model to learn spatial structural differences more effectively, a building height slicing strategy is utilised where building heights are encoded in a multi-slice representation. Each building's height is encoded in a greyscale image $g \in \{0, \ldots, 255\}$, which is linearly mapped to the physical range $[H_{min} = 6.6\ m, H_{max} = 19.8\ m]$ (3). The height range $\Delta h$ is divided into N = 12 slices (4). For each slice index $k = 0 \ldots 11$, the slice channel $S_k$ is computed (5).

$$h(i,j) = H_{min} + \frac{g(i,j)}{255} \times (H_{max} - H_{min}) \quad (3)$$

$$\Delta h = \frac{H_{max} - H_{min}}{12} \quad (4)$$

$$S_k(i,j) = \left[\frac{[h(i,j) - H_{min} - k\Delta h]}{\Delta h}\right]_0^1 \quad (5)$$

This method produces a cumulative slice encoding, where taller buildings generate non-zero activations across multiple higher-index slices. Activation values are linearly scaled in [0,1] within each slice's range. The resulting tensor provides geometric cues to the model, preserving fine-grained height information that could otherwise be lost during continuous down-sampling. The building height slicing implementation conducted on a sample urban map is illustrated in Figure 1.

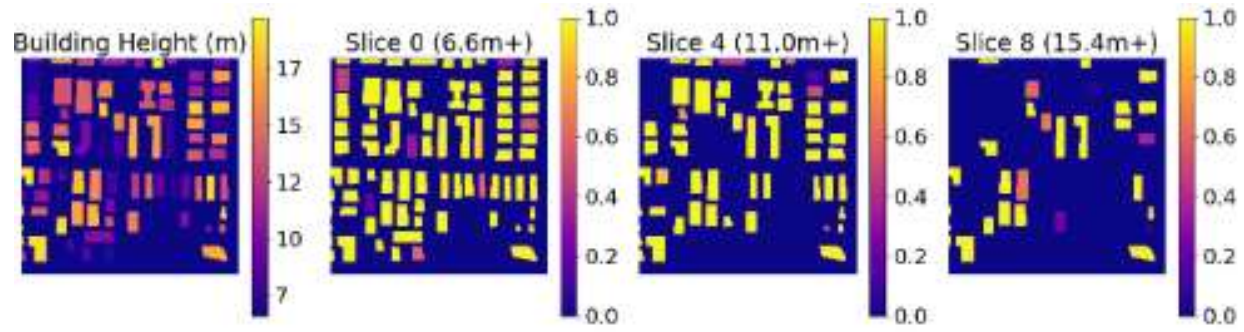

*Figure 1- Building height slicing visualisation.*

*2) Grid Anchors*

Grid Anchors (GA) encode a normalised spatial coordinate (*x*,*y*) as additional input channels, providing the model with an absolute positional reference across the urban map [9]. CNNs are translation equivariant - learned features depend on relative spatial patterns rather than absolute positions. Without positional encoding, a CNN has no explicit mechanism to distinguish absolute positions on the map. By encoding absolute spatial context, GA channels enable the model to learn distance and direction sensitive propagation patterns which increases the model accuracy and generalisation capabilities [9]. Each pixel in the map image is given a corresponding normalised (*x*,*y*) position. An example GA grid, including distance, location, and direction from the urban map centre pixel is visualised in Figure D.7 of Appendix D [12]. Adding the binary building footprint, $B(i,j) \in \{0,1\}$ and binary Tx mask, $T(i,j) \in \{0,1\}$, constructs the full sixteen-input tensor *X*, which is described by (6).

$$X \in \boldsymbol{R}^{H \times W \times C}, \qquad \boldsymbol{C} = \begin{cases} 16 \; with \; anchors \\ 14 \; without \end{cases} \tag{6}$$

*3) Data* Augmentation

Data augmentation is applied during training to improve model generalisation and robustness by exposing the model to augmented variations of the image data. Augmentation strategies are implemented as per the approach in [2] where geometric symmetries are applied to the input tensor *X*. Each epoch, one of eight possible rigid transformations is randomly applied, consisting of all combinations of four possible $90^o$ rotations $R_k, where \; k \in \{0, 1, 2, 3\}$, and a horizontal flip $H^f$, where $f \in \{0, 1\}$ is a variable representing whether to flip or not. Each training epoch (*k*, *f*) is redrawn independently to produce original samples $X'$ for the model input without storing additional images (7).

$$X' = H^f(R_k(X)) \tag{7}$$

The augmented image $X'$ preserves the physical properties of the urban map and ray-tracing simulations; while producing spatial variations that the CNN interprets as novel samples. As CNNs are not inherently rotationally invariant, this approach results in improved model generalisation, increased robustness to Tx location changes, and increases in data efficiency without requiring additional data storage. An example DA implementation is illustrated in Figure 2.

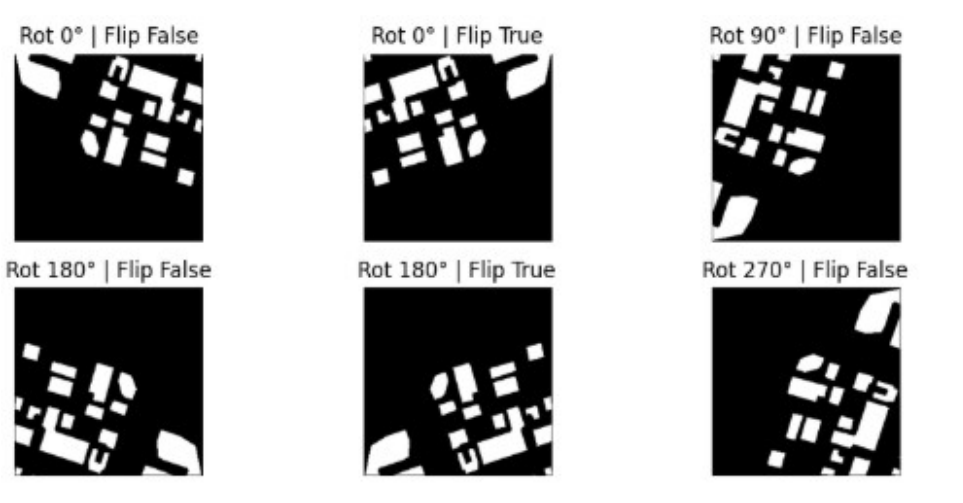

*Figure 2 - Data Augmentation example.*

## D. Encoder Selection

An ablation study was conducted to evaluate EfficientNet encoder variant scales (B0 – B7) within a U-Net architecture. The RadioMapSeer2D dataset was selected to enable faster experimentation with reduced training complexity. To isolate the effect of the EfficientNet variant, the models are trained without height slicing, GA, or DA. Each model was trained for 10 epochs using fixed hyperparameters for all experiments: learning rate of 0.001, Adam optimiser, and no explicit regularisation. Performance was evaluated using mean squared error (MSE) on the test dataset, and computational efficiency is measured in floating-point operations per second (FLOPS) to assess the accuracy and efficiency trade-off.

*1) EfficientNet Encoder*

The core building block of EfficientNet is the Mobile Inverted Bottleneck Convolution (MBConv) enhanced with squeeze and excitation (SE) modules [3] to improve channel-wise feature calibration and overall efficiency. EfficientNet employs a compound scaling method to balance network depth, width, and resolution, enabling high accuracy with fewer parameters and operations. This balanced scaling makes EfficientNet a strong candidate for spatial prediction tasks such as RME, where both precision and model efficiency are critical. In this work, the input resolution is fixed for all experiments. In EA-RMENet, the classification layer of EfficientNet is removed, and the output of the final stage is passed directly into the ASPP bottleneck. The encoder's input stem is modified to accept multi-channel tensors, 14 without GAs and 16 with GAs. Skip connection feature maps are extracted from the designated EfficientNet blocks as detailed in Table D.7 in Appendix D [11].

## E. Attention Gates (AG)

Traditional U-Net skip connections concatenate the entire encoder feature map to the corresponding decoder feature map, preserving spatial detail but potentially introducing noisy or redundant features. Attention-gated skip connections allow the decoder to selectively highlight and transmit the most relevant encoder features [5]. The mechanism is entirely learnable and enhances localisation accuracy while reducing computational load [5]. Decoder features represent a gating signal *g*, which determines the prominent components of the encoder feature map $x_E$. Both $x_E$ and *g* are linearly projected with learnable 1 x 1 convolutions, $W_x$ and $W_g$, to produce signals $\theta_x$ and $\theta_g$. The projected signals are summed and passed through a ReLu activation, followed by another 1x1 convolution and a sigmoid activation to produce attention coefficients $\alpha \in [0,1]$. The encoder features are then weighted by α to produce the attention feature map $x'$. Decoder features are spatially resized to match $x_E$ prior to computing the attention map α, ensuring dimensional consistency. In EA-RMENet, AGs are applied at all skip connections between the encoder and decoder layers. The attention gate data flow block diagram is illustrated in Figure 3.

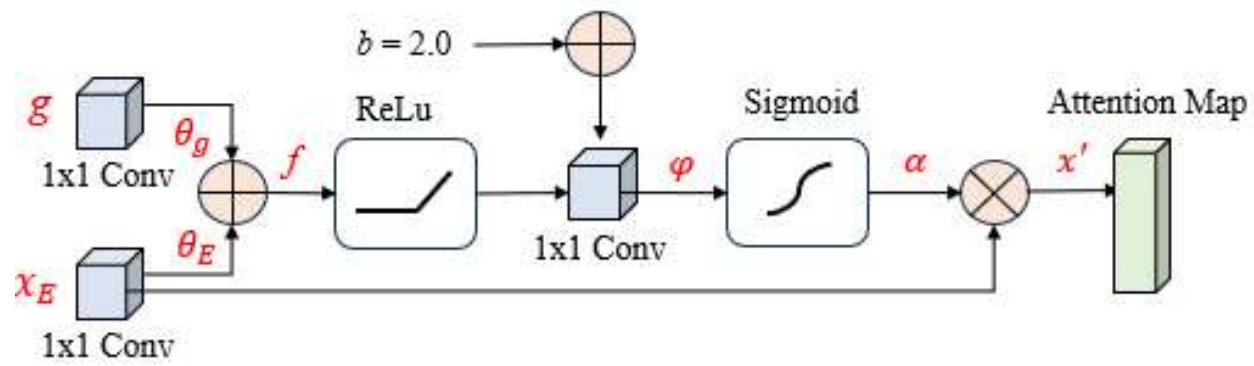

*Figure 3 - Attention gate data flow block diagram.*

## F. Atrous Spatial Pyramid Pooling (ASPP)

In CNNs, the receptive field defines the spatial range of the input influencing feature extraction [2]. For path loss modelling, relevant features span multiple scales: local effects (e.g., local obstacles) and global phenomena (e.g., shadowing, reflections, and scattering). Expanding the receptive field enables the model to capture these multi-scale propagation phenomena more effectively. Standard convolutions increase the receptive field by enlarging the kernel, which significantly increases computational load. Atrous (dilated) convolutions solve this issue by expanding the kernel's receptive field with inserted spaces between kernel elements, expanding the receptive field without significantly increasing the model parameter count [8]. The ASPP module, introduced in DeepLabV3 [8], captures multi-scale features by applying parallel atrous convolutions with varying dilation rates [8]. In EA-RMENet, a five-branch ASPP is integrated in the U-Net bottleneck to extract scale-sensitive features from the deepest encoder layer.

The first branch, $f_{GAP}$, applies global average pooling (GAP) to capture global scene context from the input feature map $x$. The pooled features are processed by a 1 x 1 convolution, batch normalisation, and ReLu activation, and then upsampled to match the input dimensions. The second branch, $f_{1x1}$, applies a 1 x 1 convolution to extract local details, followed by batch normalisation and ReLu activation. The remaining three branches ($f_{3x3}^{r=2}, f_{3x3}^{r=3}, f_{3x3}^{r=4}$), apply parallel 3 x 3 atrous convolutions with dilation rates $r = \{2,3,4\}$ to progressively increase the receptive field and capture spatial features at multiple scales. Each atrous convolution is followed by batch normalisation and ReLu activation. The outputs of all five branches are concatenated along the channel axis to produce a multi-scale feature representation. Which is passed through a fusion block ($f_{use}$) consisting of a 1 x 1 convolution, batch normalisation, and ReLu activation to generate a single fused ASPP output feature map ($y_{ASPP}$) (8) utilised in the decoder. See Figure D.10 in Appendix D [11] for the ASPP module block diagram.

$$y_{ASPP} = f_{fuse}[\, f_{GAP}(x), f_{1x1}(x), f_{3x3}^{r=2}(x), f_{3x3}^{r=3}(x), f_{3x3}^{r=4}(x)] \quad (8)$$

### G. Decoder

The decoder in EA-RMENet comprises five stages that progressively reconstruct the predicted radio map from the compressed feature representation produced by the EfficientNet encoder and the ASPP bottleneck. At each stage, the feature map is upsampled by a factor of two and fused with the corresponding encoder feature map via the AG skip connection. The fused features are subsequently processed by two consecutive 3 x 3 convolutional layers and a ReLu activation to restore detailed structural information in the output. Decoder stage spatial dimensions are outlined in Table D.8 in Appendix D [11].

## V. Results & Analysis

### A. Ablation Studies

Analysis of the EfficientNet encoder ablation study results indicates that the EfficientNetB5 offers a favourable trade-off between prediction accuracy and computational efficiency. EfficientNetB6 and B7 achieve marginally lower MSE results with a relatively significant computational cost when compared to the EfficientNetB5 results. Based on this high-level trade-off analysis, EfficientNetB5 was selected as the encoder for subsequent EA-RMENet development. The selected encoder achieved a test MSE of 0.00352 with an inference time of 0.0116 seconds per sample requiring 14.73 million FLOPS. The complete results and corresponding performance plot are provided in Table E.1 and Figure E.1 in Appendix E [12].

### B. Model Training Method

Two EA-RMENet variants are trained using different input data processing strategies to assess the impact of different input preprocessing configurations. EA-RMENet/GA incorporates grid anchors, and EA-RMENet/DA applies data during training. Both variants utilise the building height slicing method, the same holdout data split defined in Table II, and hyperparameters summarised in Table III. A cosine decay learning rate schedule is applied to progressively reduce the learning rate from $l_{rmax}$ to $l_{rmin}$ from the schedule start epoch $t_{\theta}$, to the final training epoch, enhancing generalisation and stabilising training. Weight decay (L2 regularisation) is applied to mitigate model overfitting, and the model checkpoint corresponding to the optimal validation loss is saved.

*Table II - Data splitting ratios*

| Data Split | | | |
|---|---|---|---|
| Type | Train | Validation | Testing |
| Maps | 500 | 100 | 100 |
| Samples | 40,000 | 8,000 | 8,000 |

*Table III - Model training hyperparameters.*

| Training Hyperparameters | |
|---|---|
| Batch size | 12 |
| Epochs | 40 |
| Optimiser | ADAMW |
| Initial learning rate $l_{rmax}$ | 0.001 |
| Final learning rate $l_{rmin}$ | 0.0001 |
| Start epoch | 15 |
| Weight decay | 0.00001 |

The proposed models are trained to minimise the loss function Mean Squared Error (MSE) (9) [2] where y denotes the normalised ground-truth path-loss value, $\acute{y}$ represents the predicted value, and N is the total number of image pixels. Model performance is primarily evaluated using computed Root Mean Squared Error (RMSE) (10) in the normalised range [0,1], which provides a direct correspondence to the physical path loss value in decibels (dB). Additional evaluation metrics include Mean Absolute Error (MAE) and Normalised Mean Squared Error (NMSE).

$$MSE_{loss} = \frac{1}{N}\sum_{i=1}^{N}(\, y_i - \acute{y}_i)^2 \quad (9)$$

$$RMSE_{(\acute{y},y)} = \sqrt{\frac{1}{N}\sum_{i=1}^{N}(y_i - \acute{y}_i)^2} \quad (10)$$

### C. Model Testing Method

Model performance is evaluated on the test dataset using accuracy and efficiency metrics. Accuracy is evaluated using MSE and RMSE computed as an average of all test batches across the test dataset. Efficiency is evaluated by measuring the average inference time per batch and per image sample. To assess generalisation capability, the model is tested on the ICASSP 2023 Radio Map Prediction Challenge [13] dataset, which represents a distribution shift from the training data and offers a complex test scenario. Participating in this challenge enables a direct performance comparison with state-of-the-art approaches, including PMNET [2] and RME-Net+ [14]. Model predictions for the challenge dataset are submitted for blind evaluation, and the achieved RMSE score is reported.

### D. *Training and Test Results*

Training and validation MSE loss curves are illustrated in Figure 4. The EA-RMENet/GA variant exhibited signs of overfitting to the training data at epoch 35. The corresponding validation best checkpoint was saved and selected as the EA-RMENet/GA model for further evaluation. EA-RMENet/DA demonstrated training stability over 40 epochs with a transient validation loss spike at epoch 6. This irregularity can likely be attributed to early training optimiser dynamics or the model encountered an augmented batch that generalised poorly to the validation dataset. The model training recovered in the following epoch and maintained stable convergence afterwards.

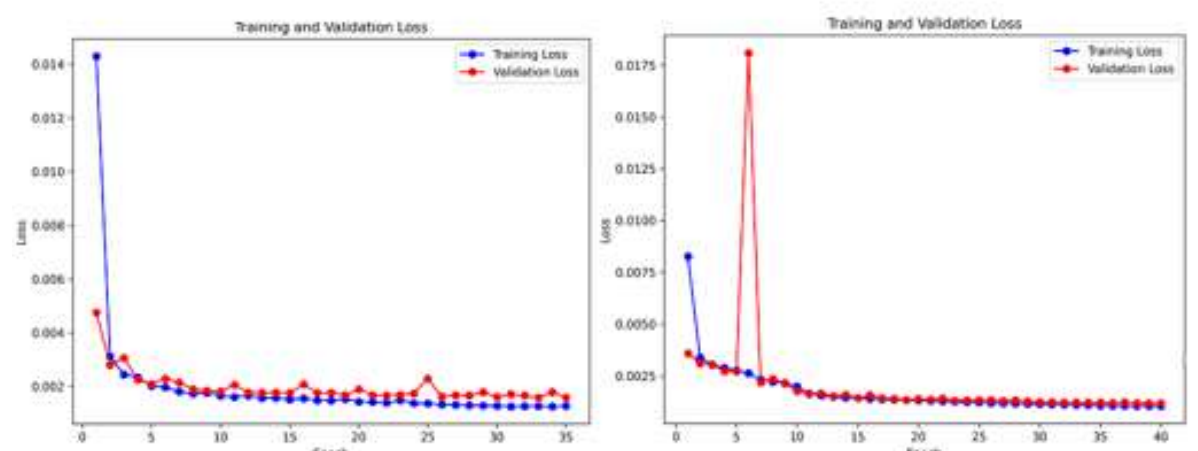

*Figure 4 - Left - EA-RMENET/GA training and validation loss plotted over 35 epochs. Right - EA-RMENET/DA training and validation loss plotted over 40 epochs.*

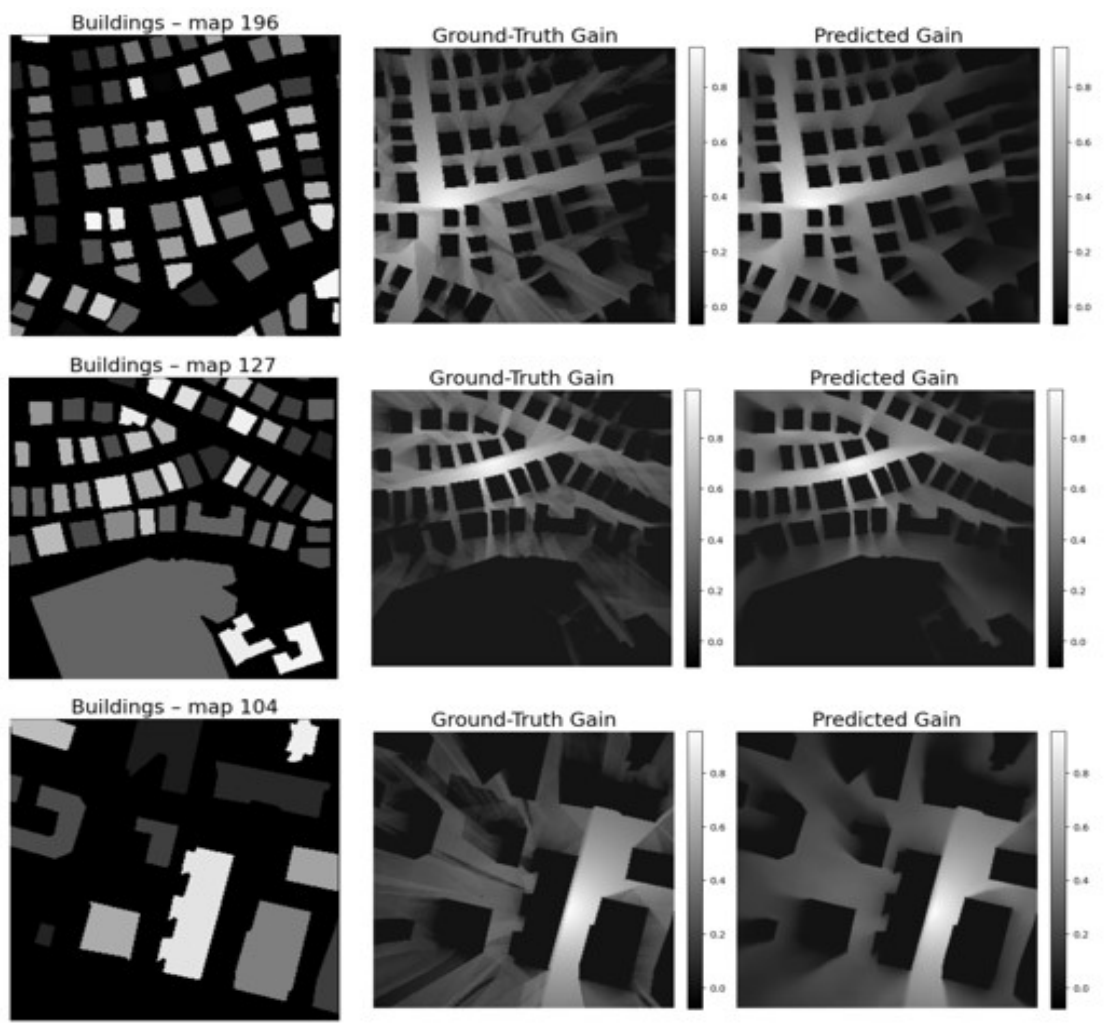

*Figure 5 - EA-RMENET/DA RME compared with ground truth.*

EA-RMENet/DA test prediction samples are illustrated in Figure 5. Both models achieved high accuracy and strong generalisation on the test dataset. EA-RMENet/GA prediction RMSE is 0.0382, indicating that explicit spatial coordinate encoding effectively supports RME by leveraging positional information. The EA-RMENet/DA variant prediction RMSE is 0.0334, outperforming EA-RMENet/GA in accuracy metrics. This demonstrates the effectiveness of DA to enhance the generalisation capability of the model. By exposing the model to transformed map variations during training, DA enhances the model's ability to learn robust features invariant to spatial transformations and enhances adaptability to unseen urban maps. EA-RMENet/DA also achieved significantly faster average inference times, due to the computational overhead associated with GA and a data pipeline re-design which is discussed in section 4.2.1 in Appendix D [11].

#### 1) *Effect of Data Augmentation & Grid Anchors*

An EA-RMENet baseline model was trained for 40 epochs without DA or GA. Baseline RMSE results are used for the evaluation of data pre-processing methods compared to the baseline results. The baseline EA-RMENet model prediction RMSE is 0.0391. Incorporating GAs reduced RMSE by 3.83% relative to baseline, and using DA reduced RMSE by 14.6%. The test results highlight the value of data augmentation for increasing generalisation and efficiency. The model characteristics that employing DA creates has been proven effective in this work for generating an accurate and efficient urban radio map prediction solution.

*Table IV - Test results for EA-RMENET/GA and EA-RMENET/DA.*

**Test Results**

| Metric | EA-RMENET/GA | EA-RMENET/DA |
|---|---|---|
| RMSE | 0.0382 | 0.0334 |
| MSE | 0.001499 | 0.001137 |
| Time/Batch | 0.4282 seconds | 0.2598 seconds |
| Time/Sample | 0.0357 seconds | 0.021663 seconds |

## VI. Performance Comparison Analysis

EA-RMENet/DA was submitted as a late entry to the IASSP 2023 Radio Map Prediction Challenge [13]. The challenge dataset includes 84 3D greyscale urban maps with 80 antenna positions per map to give a total of 6720 test samples. No corresponding ground truth images are provided. EA-RMENet/DA prediction RMSE is 0.0406 on the challenge dataset, ranking third out of five total submissions [15]. This represents an increase of 0.0072 RMSE compared to results on internal test sets, indicating a degradation in performance. Similar degradation was observed by all submissions [13]. The degradation in performance can be attributed to a distribution shift between the training and challenge datasets which includes differences in building density, more complex city map layouts, and challenging antenna positions. Increased building density in particular can amplify overlapping shadowing effects, resulting in a more challenging path loss patterns to predict [14]. The challenge leaderboard results are outlined in Table V. Sample EA-RMENET/DA challenge RM predictions are visualised in Figure 6 and Figure 7.

*Table V - ICASSP 2023 prediction challenge leaderboard.*

**ICASSP Radio Map Challenge Leaderboard**

| Method | RMSE | Date |
|---|---|---|
| REM-Net+ [15] | 0.0349 | 05/04/2024 |
| PMNet ($\frac{H}{8}\ x\ \frac{W}{8}$) [17] | 0.0383 | ICASSP 2023 SPGC |
| **EA-RMENET** | **0.0406** | **02/07/2025** |
| Agile Method [16] | 0.0451 | ICASSP 2023 SPGC |
| PPNet [16] | 0.0507 | ICASSP 2023 SPGC |

*Figure 6 - EA-RMENet ICASSP 2023 challenge prediction sample 1.*

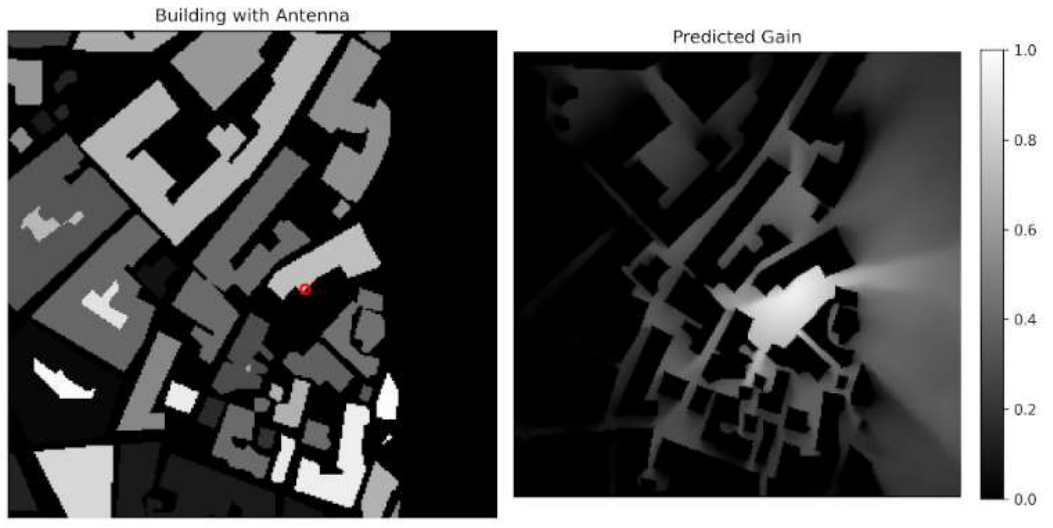

*Figure 7 - EA-RMENet ICASSP 2023 challenge prediction sample 2.*

PMNet [2] and REM-Net+ [15] report better results than EA-RMENet/DA on the challenge dataset. Both works discuss the importance of maintaining high resolution feature representations throughout the model network. PMNet demonstrated that adjusting the stride of the encoder convolutional layers to maintain a high-resolution bottleneck feature map can produce a measurable performance improvement by providing the ASPP with a wider context [16]. REMNet+ proposes a split high-resolution aggregation (SHA) structure which prioritises maintaining multi-scale spatial features throughout the network using a dedicated high-resolution pathway [15]. Further work shall investigate the integration of high-resolution feature encoding strategies into the EA_RMENET architecture. Potential strategies would include a multi-scale aggregation scheme similar to REMNET+ [15] and increasing the bottleneck resolution as demonstrated in PMNET [16].

## VII. Conclusion

This paper presented EA-RMENet, a U-Net deep learning model for urban radio map estimation (RME). The proposed model integrates an EfficientNetB5 encoder with attention-gated (AG) skip connections and an Atrous Spatial Pyramid Pooling (ASPP) bottleneck. The proposed model addresses the challenges of maintaining both accuracy and computational efficiency for RME in complex urban environments. Trained with data augmentation (DA), EA-RMENet/DA has a test RMSE of 0.0334 on the RadioMapSeer3D dataset with inference times as low as 0.022 seconds per sample. Incorporating DA into model training reduced RMSE by 14.6%. In the ICASSP 2023 Radio Map Prediction Challenge, EA-RMENet/DA ranks third in the late submission leaderboard with an RMSE of 0.0406. Results demonstrate a competitive performance that is comparable with state-of-the-art RME models. The results highlight the potential for this deep learning RME tool to be an accurate and practical alternative to traditional path loss modelling methods. Future work shall aim to improve the spatial feature learning capabilities and generalisation performance of EA-RMENet with the integration of a higher resolution encoder.

## References

[1] R. Levie and et al., "RadioUNet: Fast Radio Map Estimation With Convolutional Neural Networks," *IEEE Transactions on Wireless Communications,* vol. 20, no. 6, pp. 4001 - 4015, 2021.

[2] J.-H. Lee and et al, "PMNet: Robust Pathloss Map Prediction via Supervised Learning," in *IEEE Global Communications Conference*, Kuala Lumpur , 2023.

[3] M. Tan and et al., "EfficientNet: Rethinking Model Scaling for Convolutional Neural Networks," in *Proceedings of the 36th International Conference on Machine Learning, PMLR*, Long Beach, 2019.

[4] C. Yapar and et al., "Dataset of Pathloss and ToA Radio Maps with Localization Application," arXiv:2212.11777v4, 2022.

[5] O. Oktay and et. al., "Attention U-Net: Learning Where to Look for the Pancreas," arXiv:1804.03999v3, 2018.

[6] Z. Fu and et al., "Transfer Learning and Double U-Net Empowered Wave Propagation Model in Complex Indoor Environments," *IEEE Transactions on Antennas and Propagation,* 2025.

[7] B. Baheti and et al., "Eff-UNet: A Novel Architecture for Semantic Segmentation in Unstructured Environment," in *IEEE/CVF Conference on Computer Vision and Pattern Recognition Workshops (CVPRW)*, Seattle, 2020.

[8] L.-C. Chen and et al, "DeepLab: Semantic Image Segmentation with Deep Convolutional Nets, Atrous Convolution, and Fully Connected CRFs," *IEEE Transactions on Pattern Analysis and Machine Intelligence,* vol. 99, 2016.

[9] Y. Tian and et al., "Transformer based Radio Map Prediction Model for Dense Urban Environments," in *13th International Symposium on Antennas, Propagation and EM Theory (ISAPE)*, Zhuhai, 2021.

[10] M. Diab and et al, "A Deep Learning Approach for Automatic Liver Segmentation: Attention U-Net with ASPP and CBAM Integration," in *2025 6th International Conference on Bio-engineering for Smart Technologies (BioSMART)*, Paris, 2025.

[11] J. O'Shea, "Appendix D," in *Radio Map Prediction in Urban Environments*, Dublin, DCU, 2025.

[12] J. O'Shea, "Appendix E," in *Radio Map Prediction in Urban Environments*, Dublin, DCU, 2025.

[13] C. Yapar and et al., "Overview of the First Pathloss Radio Map Prediction Challenge," *IEEE Open Journal of Signal Processing,* vol. 5, pp. 948-963, 2024.

[14] Q. Chen and et al., "REM-Net+: 3D Radio Environment Map Construction Guided by Radio Propagation Model," *TexhRix,* 29 07 2024.

[15] radiomapchallenge, "The First Pathloss Radio Map Prediction Challenge," ICASSP, 2023. [Online]. Available: https://radiomapchallenge.github.io/results.html. [Accessed 10 07 2025].

[16] J.-H. Lee and et al., "PMNet: Large-Scale Channel Prediction System for ICASSP 2023 First Pathloss Radio Map Prediction Challenge," in *IEEE ICASSP 2023*, Rhodes Island, 2023.